\def\year{2022}\relax
\documentclass[letterpaper]{article} 
\usepackage{aaai20}  
\usepackage{times}  
\usepackage{helvet} 
\usepackage{courier}  
\usepackage[hyphens]{url}  
\usepackage{graphicx} 
\usepackage{graphicx}  
\usepackage{pgfplots}
\pgfplotsset{compat=newest}
\usetikzlibrary{matrix}
\usepackage{amsmath,amssymb,amsfonts}
\usepackage{subfig}
\usepackage{latexsym}
\usepackage{url}
\usepackage{algorithm}
\usepackage{algpseudocode}
\usepackage{graphicx}
\usepackage{textcomp}
\usepackage{xcolor}

\newtheorem{definition}{Definition}

\nocopyright

\title{Leveraging Planning Landmarks for Hybrid Online Goal Recognition}
\author{Nils Wilken,\textsuperscript{\rm 1} Lea Cohausz,\textsuperscript{\rm 2} Johannes Schaum,\textsuperscript{\rm 1} Stefan Lüdtke,\textsuperscript{\rm 1} Christian Bartelt,\textsuperscript{\rm 1}  \\ \Large \textbf{Heiner Stuckenschmidt\textsuperscript{\rm 2}}
\\
\textsuperscript{\rm 1}Institute for Enterprise Systems, University of Mannheim, Mannheim, Germany\\
\textsuperscript{\rm 2}Data and Web Science Group, University of Mannheim, Mannheim Germany\\
\{bartelt, luedtke, wilken\}@es.uni-mannheim.de, jschaum@mail.uni-mannheim.de,\\ \{lea, heiner\}@informatik.uni-mannheim.de
}

\begin{document}

\maketitle

\begin{abstract}
Goal recognition is an important problem in many application domains (e.g., pervasive computing, intrusion detection, computer games, etc.).
In many application scenarios it is important that goal recognition algorithms can recognize goals of an observed agent as fast as possible and with minimal domain knowledge.
Hence, in this paper, we propose a hybrid method for online goal recognition that combines a symbolic planning landmark based approach and a data-driven goal recognition approach and evaluate it in a real-world cooking scenario.
The empirical results show that the proposed method is not only significantly more efficient in terms of computation time than the state-of-the-art but also improves goal recognition performance.
Furthermore, we show that the utilized planning landmark based approach, which was so far only evaluated on artificial benchmark domains, achieves also good recognition performance when applied to a real-world cooking scenario.
\end{abstract}

\section{Introduction}
Goal recognition is the task of recognizing the goal(s) of an observed agent from a possibly incomplete sequence of actions executed by the observed agent.
This task is relevant in many real-world application domains like crime detection \cite{geib2001plan}, pervasive computing \cite{wilken2021hybrid}, \cite{geib2002problems}, or traffic monitoring \cite{pynadath1995accounting}.
State-of-the-art goal recognition systems often rely on the principle of Plan Recognition As Planning (PRAP) and hence, utilize classical planning systems to solve the goal recognition problem \cite{ramirez2009plan}, \cite{ramirez2010probabilistic}, \cite{sohrabi2016revisited}, \cite{amado2018lstm}.
Recently, a hybrid goal recognition method that combines a PRAP approach and a data-driven approach for goal recognition \cite{wilken2021hybrid} was proposed and it was shown that this hybrid method is able to outperform both, the single PRAP approach and the single data-driven approach.

Nevertheless, a major obstacle of the proposed hybrid method is that it requires large amounts of computation time to solve goal recognition problem instances of real-world complexity (in the range of hours).
This is a significant problem when this approach should be used in a near real-time online recognition setting in which fast goal recognition is key.
Moreover, as the utilized data-driven approach is, once it is trained, able to perform goal recognition in near real-time, the planning based approach is the main bottleneck of the hybrid method.
As a solution, we propose to use an approach that based on so called \textit{planning landmarks}, which are widely used in classical planning research to structure the search during the planning process \cite{hoffmann2004ordered}, instead.
This landmark based approach was first introduced by \cite{pereira2020landmark} and requires significantly less computation time to solve goal recognition problem instances and hence, reduces the computation time of the entire hybrid method.
More explicitly, the contributions of this paper are:
\begin{itemize}
    \item In Section \ref{sec:landmarkBasedGoalRecognition}, we discuss some extensions and adaptions of the existing landmark based approach that are necessary so that it can be applied to more complex planning domains that are not restricted to the STRIPS part of PDDL which is required by the domain used in the evaluation.
    \item Also in Section \ref{sec:landmarkBasedGoalRecognition}, we propose to ignore trivial landmarks for goal recognition as this is expected to improve goal recognition performance.
    \item In Section \ref{sec:leveragingPlanningLandmarks}, we discuss how the planning landmark based approach can be used in a hybrid goal recognition approach.
    \item Finally, we empirically evaluate the proposed changes and extensions to the landmark based method, which was so far only evaluated on standard benchmarks from the goal recognition literature, and the landmark based hybrid goal recognition method on a real world cooking scenario in Section \ref{sec:evaluation}.
\end{itemize}
The empirical evaluation shows that the landmark based approach requires dramatically less computation time than the so far used planning based approach, which, as a consequence, also reduces the computation time required by a planning landmark based hybrid goal recognition method.

\section{Problem Definition}
\label{sec:problemDefinition}
In this work, we investigate a possible solution method for the online probabilistic goal recognition problem in a real-world scenario.
Before we formally define the online probabilistic goal recognition problem, we start by defining the probabilistic goal recognition problem.
\begin{definition}[Probabilistic Goal Recognition]
\label{def:probabilisticGoalRecognition}
\textit{Probabilistic goal recognition} is the problem of inferring a probability distribution over a set of intended goals of an observed agent, given a possibly incomplete sequence of observed actions and a domain model $D = \langle F, s_0, A \rangle$, where $F$ is a set of facts, $s_0$ is the initial state, and $A$ is a set of actions.
More details on the background of symbolic planning domains is explained in Section \ref{subsec:symbolicPlanning}.
More formally, the aim of goal recognition approaches is to find a \textit{posterior} probability distribution $P(G|\pmb{o})$ for all goals $g \in G$ given a sequence of observed actions $\pmb{o}$.
\end{definition}
The online probabilistic goal recognition problem is an extension to the previously defined probabilistic goal recognition problem that additionally introduces the concept of time:
\begin{definition}[Online Probabilistic Goal Recognition]
\label{def:onlineGoalRecognition}
We define \textit{online probabilistic goal recognition} as a special variant of the \textit{probabilistic goal recognition} problem.
In online goal recognition, we assume that the observation sequence $\pmb{o}$ is revealed incrementally.
More explicitly, we introduce the notion of time $t \in \{0, \dots, T\}$, where $T = |\pmb{o}|$.
For every value of $t$, one probabilistic goal recognition problem $R(t)$ can be induced as $R(t) = \langle D, G, \pmb{o_t}, P(G) \rangle$ where $\pmb{o_t} = \{o_i | 1 \leq i \leq t, o_i \in \pmb{o}\}$, $P(G)$ is a prior distribution over the set of goals.
A solution to the online probabilistic goal recognition problem are the conditional probabilities $P_t(G=g|\pmb{o_t}); \forall g \in G, t \in [0,T]$.
\end{definition}
Hence, solving an online probabilistic goal recognition problem is similar to solving a sequence of probabilistic goal recognition problems, where for each problem instance in this sequence the utilized observation sequence is extended by the next observed action.

\section{Background}
\label{sec:planningAndLandmarks}
In the context of classical planning systems, planning landmarks are usually utilized to guide the heuristic search through the search space that is induced by a planning problem \cite{hoffmann2004ordered}.
However, in this work, we study the utilization of a planning landmark based goal recognition approach (PLR), which was originally introduced by \cite{pereira2020landmark}, in a hybrid goal recognition method.
The basic idea of PLR is to use the structural information that can be derived from planning landmarks, which can be - informally - seen as way-points that have to be passed by every path to a possible goal.
Hence, when we have observed that such way-points were passed recently by an agent, this indicates that the agent currently might follow a path to the goal(s) for which the observed way-point is a landmark.

\subsection{Classical Planning}
\label{subsec:symbolicPlanning}
Classical planning is usually based on a model of the planning domain that defines possible actions, their preconditions, and effects on the domain.
More formally, in this work, we define a (STRIPS) planning problem as follows:
\begin{definition}[Planning Problem]
A Planning Problem is a tuple $P = \langle F, s_0, A, g \rangle$ where $F$ is a set of facts, $s_0 \subseteq F$ and g $\subseteq F$ are the initial state and a goal and $A$ is a set of actions with Preconditions $Pre(a) \subseteq F$ and lists of facts $Add(a) \subseteq F$ and $Del(a) \subseteq F$ that describe the effects of an action $a$ in terms of facts that are added and deleted from the current state.
Actions have a non-negative cost $c(a)$.
A state is a subset of $F$.
A goal state is a state $s$ with $s \supseteq g$.
An action $a$ is applicable in a state $s$ if and only if $Pre(a) \subseteq s$.
Applying an action $a$ in a state $s$ leads to a new state $s' = (s \cup Add(a) \setminus Del(a))$.
A solution for a planning problem (i.e., a plan) is a sequence of applicable actions $\pi = a_1, \cdots a_n$ that transforms the initial state into a goal state.
The cost of a plan is defined as $c(\pi) = \sum \limits_i c(a_i)$.
A plan is optimal if the cost of the plan is minimal.  
\end{definition}

\subsection{Extracting Planning Landmarks}
\label{subsec:extractingPlanningLandmarks}
Planning landmarks are typically defined as facts that must hold or actions that must be executed at some point during the execution of a valid plan starting at $s_0$ that achieves the goal $g$ \cite{hoffmann2004ordered}.
In this work, we only focus on \textit{fact landmarks}.
More precisely, following \cite{hoffmann2004ordered}, we define fact landmarks as follows:
\begin{definition}[Fact Landmark]
\label{def:planningLandmark}
Given a planning problem $P = \langle F, s_0, A, g \rangle$, a fact $f \in F$ is a fact landmark if for all plans $\pi = \langle a_1, \dots, a_n \rangle$ that reach $g$: $\exists s_i: f \in s_i; 0 \leq i \leq n$, where $s_i$ is the planning state that is reached by applying action $a_i$ to state $s_{i-1}$. 
\end{definition}
\cite{hoffmann2004ordered} further divide this set of fact landmarks into \textit{trivial} and \textit{non-trivial} landmarks.
They consider all landmarks that are either contained in the initial state (i.e., $f \in s_0$) or are part of the goal description (i.e., $f \in g$) as trivial landmarks because they are trivially given by the planning problem definition.
All other landmarks are considered to be non-trivial.
\begin{figure}[htbp]
    \centering
    \includegraphics[width=0.75\linewidth,height=4.5cm]{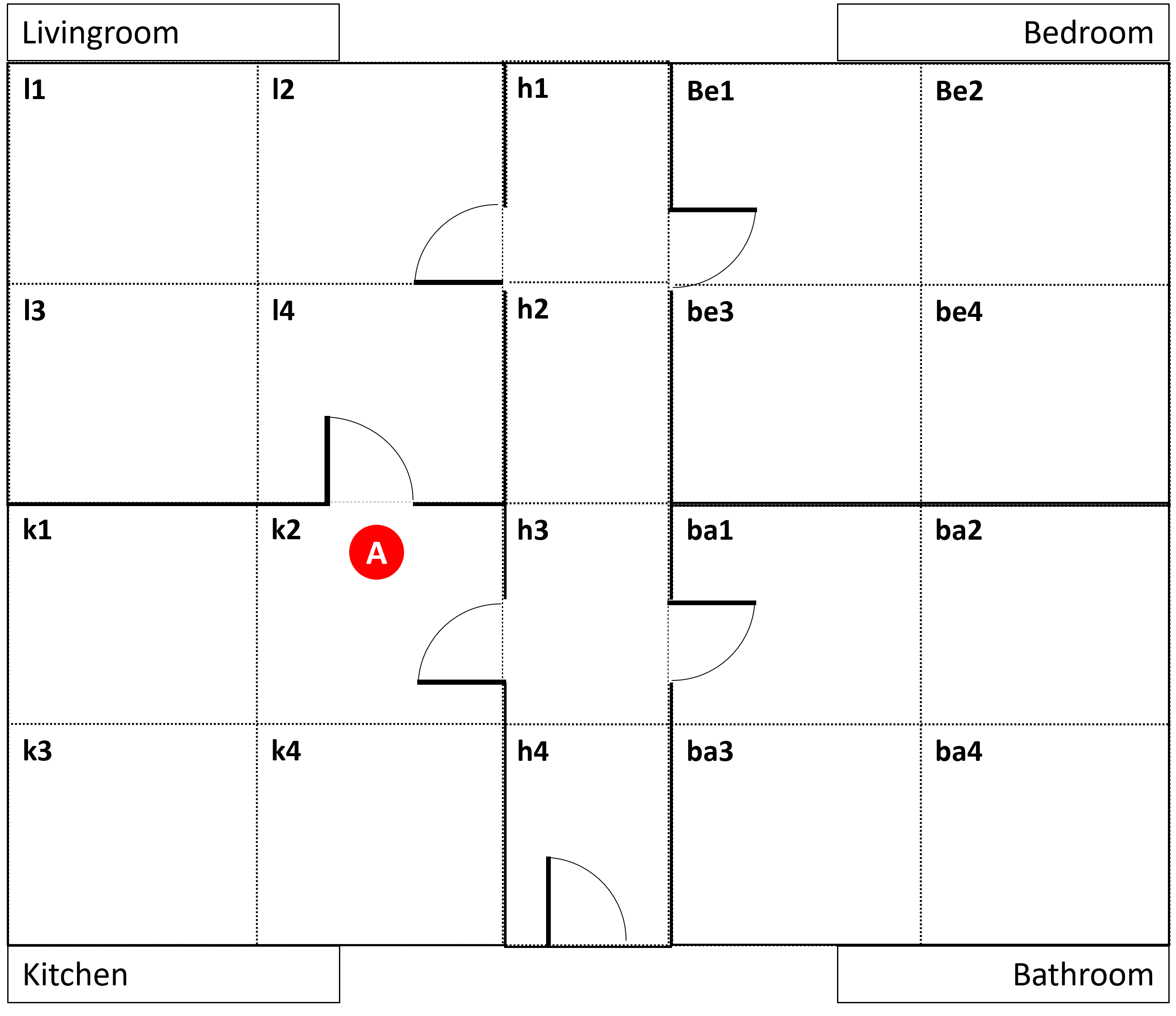}
    \caption{Exemplary Smart Home Layout.}
    \label{fig:LandmarkExample}
\end{figure}
As an example, consider the smart home scenario depicted in Figure \ref{fig:LandmarkExample}.
For this example, we assume, that the corresponding planning domain uses a predicate \textit{(is-at ?x)} to describe the current position of the agent (e.g., in the depicted state the grounded fact \textit{(is-at k2)} is true).
For this example, one potential goal of the agent is defined as $g = \{$\textit{(is-at ba3)}$\}$.
When we assume that the agent can carry out movement actions from one cell to any adjacent cell, then the facts \textit{(is-at h3)} and \textit{(is-at ba1)} would be \textit{non-trivial} fact landmarks because these cells have to be visited by every valid path from the initial position k2 to the goal position ba3 but are not part of the initial state or the goal.
Moreover, \textit{(is-at k2)} and \textit{(is-at ba3)} would be \textit{trivial} landmarks because they also have to be true on every valid path but they are given in the initial state and the goal definition respectively.

To extract landmarks, we use an algorithm that was proposed by \cite{hoffmann2004ordered}.
However, the original version of the extraction algorithm is designed to only work on the STRIPS \cite{fikes1971strips} subset of the Planning Domain Definition Language (PDDL) \cite{mcdermott1998pddl}.
As the real-world planning domain that is used in the evaluation requires some more advanced parts of PDDL, we had to slightly adjust the original algorithm.
The original algorithm \cite{hoffmann2004ordered} generates a directed landmarks generation graph (LGG).
When the algorithm is finished, all nodes in the LGG represent the detected landmarks and the edges in the graph represent an ordering relation between the extracted landmarks.
Landmarks at the tail of an edge have to be achieved before the landmark at the head of this edge can be achieved.
The original algorithm has two phases.
First, \textit{landmark candidates} are generated and added to the LGG.
Then, in a second step, all landmark candidates are evaluated to check whether they are actually landmarks.

\textit{Generation of Landmark Candidates.}
Landmark candidates are generated using a back-chaining procedure that loops backwards through the layers of an Relaxed Planning Graph (RPG) \cite{hoffmann2001ff}.
An RPG is a relaxed representation of a planning graph that ignores all delete effects.
An RPG is structured into two alternating types of layers (i.e., fact layers and action layers).
Fact layers contain all facts that might be true after the execution of at least $l$ action steps, where $l$ is the level of the corresponding fact layer.
Action layers have the same properties unless they contain all actions that might be executed after at least $l$ action steps.
The back-chaining procedure starts with considering all facts that are part of $g$, which are all part of the last layer of the RPG \textit{if} the considered planning problem is solvable.
For each of these facts $f$, the intersection of preconditions for all actions $a$ that achieve $f$ (i.e., $f \in Add(a)$) and that are part of the previous action layer is calculated.
All facts that are part of this intersection are then considered to be landmark candidates and added to the LGG because they are a precondition of all actions that achieve $f$ and hence, are potentially relevant to achieve $f$.
After this procedure is completed for all facts that are part of the goal description, the back-chaining process continues with looping through all landmark candidates that were added during the previous iteration and performing the same check.
Note that the extracted set of landmark candidates that is generated by this algorithm is not exhaustive \cite{hoffmann2004ordered}.

Nevertheless, as already mentioned, this algorithm is designed to only work for planning problems that solely use the STRIPS subset of PDDL.
The main differences, relevant in the context of this landmark extraction algorithm are that in STRIPS problems, action preconditions can, by definition, only be a conjunction of facts.
Hence, in a STRIPS problem it is quite straightforward to determine the intersection of several action preconditions.
Once more complex constructs (e.g., disjunctions, existance quantors, etc.) are used in the preconditions, computing an intersection of entire preconditions is not easily possible anymore.
As a solution, we decided to treat also more complex preconditions as if they were a conjunction of all facts that are part of it.
Hence, we add the intersection of all facts that are part of the preconditions, which is basically the same as \cite{hoffmann2004ordered} do.
However, as a consequence of this procedure, the ordering information that is generated by the original algorithm is lost as not all facts in the considered preconditions have to be necessarily true to execute an action.

\textit{Evaluation of Landmark Candidates.}
After the generation of landmark candidates, each generated candidate is evaluated to check whether it is actually a landmark or not.
This check is done through the generation of an RPG for a slightly modified planning problem which is determined by removing all actions that have the currently considered landmark candidate $lc$ as an effect (i.e., $\{a\in A|lc \in Add(a)\}$).
In the case that the modified planning problem is not solvable anymore, the examined landmark candidate is proofed to be a landmark.
This is because in this case there is no alternative action sequence, which does not contain an action that has $lc$ as an effect, that reaches the goal and hence, $lc$ has to be true in all plans that achieve this goal.

\section{Landmark Based Goal Recognition}
\label{sec:landmarkBasedGoalRecognition}
To perform goal recognition based on the information that can be gained from extracted planning landmarks, we use an adjusted version of the approach that was proposed by \cite{pereira2020landmark}.
The main reasons for the adjustments are that we think that considering trivial landmarks provides no additional benefit to solve the goal recognition problem and might even have a negative impact on the recognition performance and, in addition, that the domain used for the evaluation does not only use the STRIPS subset of PDDL.

\subsection{Computing Achieved Landmarks}
\label{subsec:computingAchievedLandmarks}
The two heuristics to estimate $P(G|O)$ both reason over the set of landmarks that were already achieved by a given observation sequence $\pmb{o}$ for each goal $g \in G$, which is refered to as $AL_g$.
To determine the set of achieved landmarks for each goal, we use the algorithm which is described in Algorithm \ref{alg:computeAchievedLandmarks}.
This algorithm is inspired by the original algorithm proposed by \cite{pereira2020landmark}.
\begin{algorithm}
\caption{Compute achieved landmarks for each goal.}
\label{alg:computeAchievedLandmarks}
\textbf{Input: } \textit{$I$ initial state, $G$ set of candidate goals, $\pmb{o}$ observations, and a set of extracted landmarks $L_g$ for each goal $g \in G$}. \\
\textbf{Output: } \textit{A mapping $M_G$ between each goal $g \in G$ and the respective set of achieved landmarks $AL_g$.}
\begin{algorithmic}[1]
\Function{Compute Achieved Landmarks}{$I$, $G$, $\pmb{o}$, $L_G$}
\State $M_G \leftarrow \langle \rangle$
\ForAll{$g \in G$}
    \State $L_g \leftarrow$ all fact landmarks from $L_g$ s.t.
    \State \hspace*{1cm}$\forall l \in L_g: l \notin I$
    \State $L \leftarrow \emptyset$
    \State $AL_g \leftarrow \emptyset$
    \ForAll{$o \in \pmb{o}$}
        \State $L \leftarrow \{l \in L_g|l \in Pre(o) \cup Add(o) \wedge l \notin L\}$
        \State $AL_g \leftarrow AL_g \cup L$
    \EndFor
    \State $M_G(g) \leftarrow AL_g$
\EndFor
\State \Return $M_G$
\EndFunction
\end{algorithmic}
\end{algorithm}
Nevertheless, it substantially differs in two points.
First, it is not able to consider the predecessor landmarks for each landmark that was detected to be achieved by the given observations.
The reason for this is that ordering information between landmarks would be necessary to do this.
However, such information are not generated by the adjusted landmark extraction procedure.
As a consequence, the adjusted algorithm to compute achieved landmarks will very likely have more difficulties dealing with missing observations compared to the original algorithm.
Second, in contrast to the original algorithm, Algorithm \ref{alg:computeAchievedLandmarks} does not consider trivial landmarks that are part of the initial state to be actually \textit{achieved} by the given observation sequence $\pmb{o}$.
Instead, these landmarks are simply ignored during the goal recognition process.
We propose this adjustment because we think that landmarks which are part of the initial state do not provide any valuable information for goal recognition but might potentially even have a misleading effect.
This is because using initial state landmarks for goal recognition in fact means that information which is not derived from the observed agent behaviour is used for recognition.
Although, the initial state (and hence all initial state landmarks) are the same for all possible goals, due to how the two recognition heuristics and the utilized planning domain are defined, using initial state landmarks introduces a bias towards considering goals with smaller numbers of non-trivial landmarks as more probable.
As a consequence, the goal(s) that have the largest fraction of their landmarks in the initial state are considered to be most probable in the initial state.
However, this is only caused by how the domain and goal descriptions are defined and not by actually observed agent behaviour.

\subsection{Estimating Goal Probabilities}
\label{subsec:estimatingGoalProbabilities}
To estimate the goal probabilities from the sets of all extracted landmarks (i.e., $L_g$) and landmarks already achieved by $\pmb{o}$ (i.e., $AL_g$) for each $g \in G$, we use slightly adjusted versions of the heuristics introduced by \cite{pereira2020landmark}.
One heuristic considers the percentage of completion in terms of the fraction of all landmarks that were already identified as achieved by the given observation sequence.
The second heuristic computes a uniqueness score for each landmark and uses these scores for the computation of the heuristic scores.

\textit{Goal Completion Heuristic.}
The original version of this heuristic estimates the completion of an entire goal as the average of completion percentages of the sub-goals (i.e., all facts $sg \in g$, where $g \in G$) of a goal.
More precisely, the original heuristic is computed as follows \cite{pereira2020landmark}:
\begin{equation}
\label{eq:originalCompletionHeuristic}
    h_{gc}(g, AL_g, L_g) = \Bigg( \frac{\sum_{sg \in g}{\frac{|AL_{sg}|}{|L_{sg}|}}}{|g|} \Bigg)
\end{equation}
However, to which of the sub-goals each of the identified achieved landmarks contributes can again only be determined if ordering information between the landmarks is available.
Hence, as the landmark extraction method that is used in this work does not generate such information, the completion was slightly adjusted to be computed as:
\begin{equation}
\label{eq:adjustedCompletionHeuristic}
    h_{gc}(g, AL_g, L_g) = \Bigg( \frac{|AL_g|}{|L_g|} \Bigg)
\end{equation}
This adjustment, in some cases, has a significant impact on the resulting heuristic scores.
For example, consider the case that $g = \{sg_0, sg_1, sg_2, sg_3, sg_4\}$, $|L_{sg_i}| = 1$ and $|AL_{sg_i}| = 1$, $\forall sg_{i} \in g; 0 \leq i \leq 3$, $|AL_{sg_4}| = 0$, and $|L_{sg_4}| = 30$.
In this case, the result of Equation \ref{eq:originalCompletionHeuristic} would be $4/5$, whereas the result of Equation \ref{eq:adjustedCompletionHeuristic} would be $4/34$.
Thus, the more unevenly the number of landmarks is distributed over the sub-goals, the larger the difference between the original heuristic calculation and the adjusted calculation becomes.
Nevertheless, it is not fully clear which of the two options achieves better goal recognition performance.

\textit{Landmark Uniqueness Heuristic.}
The second heuristic that was proposed by \cite{pereira2020landmark} does not only consider the percentage of completion of a goal in terms of achieved landmarks but also considers the uniqueness of the landmarks.
The intuition behind this heuristic is that it is quite common that several goals share a common set of fact landmarks.
Hence, landmarks that are only landmarks of a small set of potential goals (i.e., landmarks that are more unique) provide us with more information regarding the most probable goal than landmarks that are landmarks for a larger set of goals.
For this heuristic, \textit{landmark uniqueness} is defined as the inverse frequency of a landmark among the found sets of landmarks for all potential goals.
More formally the landmark uniqueness is computed as follows \cite{pereira2020landmark}:
\begin{equation}
    \label{eq:landmarkUniquenessScore}
    L_{uniq}(l,L_G) = \Bigg(\frac{1}{\sum_{L_g \in L_G}{|\{l|l \in L_g\}|}}\Bigg)
\end{equation}
Following this, the uniqueness heuristic score is computed as:
\begin{equation}
    \label{eq:uniquenessHeuristic}
    h_{uniq}(g, AL_g, L_g, L_G) = \Bigg(\frac{\sum_{al \in AL_g}{L_{uniq}(al, L_G)}}{\sum_{l \in L_g}{L_{uniq}(l, L_G)}}\Bigg)
\end{equation}

To determine the set of most probable goals, for both heuristics, we calculate the heuristic values for all potential goals and then consider the set of goals that are assigned with the highest heuristic score as most probable goals.

\section{Hybrid Landmark Based Goal Recognition}
\label{sec:leveragingPlanningLandmarks}
Recently, a hybrid method for goal recognition was proposed by \cite{wilken2021hybrid} to overcome some identified shortcomings of purely symbolic methods.
However, as already mentioned, one major shortcoming of the proposed hybrid method is that the so far used planning based method, which was first proposed by \cite{ramirez2010probabilistic} (we will refer to this approach as ``RG'' from here on), requires a tremendous amount of computation time.
This is a major issue, especially when these methods should be applied to online goal recognition scenarios of real-world complexity.
To overcome this shortcoming, we propose to use the PLR method in the hybrid method.
This significantly reduces the required computation time, as the PLR method no longer requires to actually solve several planning problems for each step in time for an online goal recognition problem (see Definition \ref{def:onlineGoalRecognition}) but only has to extract the fact landmarks once at time step $t = 0$ and afterwards, only has to update the sets of achieved landmarks for each goal.

To combine the estimate of the PLR approach with an estimate of a data-driven method to obtain a hybrid estimate, we follow the approach of \cite{wilken2021hybrid}.
They have investigated two different combination schemes to combine the goal probability estimates of the RG method and a \textit{Bayesian Network} (BN) model.
In this work we only use the weighted sum combination scheme, as it was found that it achieves better performance.

\subsection{Probabilistic Goal Recognition Model}
\label{subsec:probabilisticGoalRecModel}
To model probabilistic knowledge about the environment, we use a \textit{Bayesian Network} (BN) model with the same topology as in \cite{wilken2021hybrid}.
Essentially, the topology of the used BN equals a Naive Bayes Model (NBM) that treats the goal recognition problem as a classification problem.
The NBM, in general, has one random variable for each observable planning fact (i.e., $F_i$) and another random variable that represents the possible goals.
Hence, it estimates the probability of a goal $g$ given an observation sequence $\pmb{o}$ as follows:
\begin{equation}
P(\pmb{o}|g) = P(F_1, \dots, F_n|g) = \prod_{i=1}^{n}{P(F_i|g)}
\end{equation}
Where $F_1, \dots, F_n$ is the set of observable planning facts that are defined in the planning domain.

\section{Evaluation}
\label{sec:evaluation}
To evaluate the performance and efficiency of the adjusted methods discussed in the previous sections, we conducted several empirical experiments on a real-world data set (i.e., CMU Grand Kitchen Challenge \footnote{\url{http://kitchen.cs.cmu.edu/index.php}}).
More precisely, the goals of the evaluation are:
\begin{itemize}
    \item Show that ignoring \textit{trivial} landmarks that are part of the initial state during the goal recognition process improves the recognition performance.
    \item Show that the PLR method achieves significantly better goal recognition performance than the RG method when applied to a goal recognition scenario of real-world complexity.
    \item Show that the PLR approach, and in consequence also the hybrid recognition approach, requires significantly less computation time than the RG method.
    \item Show that a PLR and NBM based hybrid goal recognition method outperforms both single approaches.
\end{itemize}

\subsection{Experimental Setup}
We conducted several empirical experiments with the proposed adjusted hybrid method as well as the PLR method, the RG method, and the NBM on a real-world dataset to achieve the previously mentioned evaluation goals.
In all experiments of this evaluation, the online goal recognition problem is considered (see Definition \ref{def:onlineGoalRecognition}).
All experiments of this evaluation were carried out on machines that have 24 cores with 2.60GHz and at least 386GB RAM.
The remainder of this subsection describes the utilized dataset and different experimental setups.

\textit{Dataset.}
As a real-world data set, we used the CMU-MMAC Kitchen Dataset \cite{torre2009data}.
This dataset contains data from different sources (e.g., video, motion capture, etc.) that were recorded by observing different persons while cooking one out of five different recipes.
We will consider reaching the end of the cooking process for each of the recipes as possible goals.
We first had to transform the existing ``raw'' data into a suitable format for our purpose.
As a starting point for this transformation, we used the results of a semantic annotation project at the University of Rostock \cite{yordanova2018data}.
In this project, planning domains in PDDL format and annotated observation sequences were created for three of the five recipes (i.e., brownies, eggs, and sandwich).
In addition, we created annotations for the remaining two recipes (i.e., pizza and salad).
Consequently, the set of possible goals is defined as $G_{CMU} = \{brownies,eggs,sandwich,pizza,salad\}$.
In total, the dataset contains 148 full observation sequences.

\textit{PLR and RG Setup.}
We implemented the PLR and RG approaches in Java using the PPMAJAL \footnote{\url{https://gitlab.com/enricos83/PPMAJAL-Expressive-PDDL-Java-Library}} library for PDDL related functionalities.
To solve the planning problems for the RG approach, we used the MetricFF \cite{hoffmann2003metric} planner.
MetricFF is a satisficing planner that supports metric facts, which is required by the planning domain utilized in this evaluation.
We use the planner in a ``greedy'' mode which means that the planner always returns the first found plan as the solution.
In addition, we used a timeout of 360 seconds.
Problems for which no solution was found after this time are considered to be unsolvable.
During the experiments, this happened regularly especially for the planning problems that require to determine a plan that does not fulfill the observation sequence.
Moreover, we assume equal costs of one for all actions and set the $\beta$ parameter of the RG approach to 1.
The $\beta$ parameter is a measure of the assumed rationality of the observed agent, where a value of 0 represents completely irrational and a value of 1 completely rational.

\textit{NBM Setup.}
The NBM used for this evaluation contains one random variable (RV) for the possible user goals and one RV for each fact in the planning domain.
We define the sample space of RV $X_{goal}$ as $S_{X_{goal}} = G$.
For all other RVs $X_f$, we assume that the sample space $S_{X_f}$ is defined as $S_{X_f} = \{true,false\}$.
This corresponds to the nature of planning facts.

\textit{Combining PLR and NBM Methods.}
For the weighted sum, we compute the weight for the NBM as \(w_{NBM}(n) = \frac{a}{1 + e^{-b(n-c)}}\), where $a$, $b$, and $c$ are fitting parameters.
For this evaluation, we set the parameters to $a=0.7$, $b=0.45$, and $c=11.5$.
The weight for the PLR approaches is then calculated as \(w_{PLR} = 1 - w_{NBM}(n)\).

\textit{K-Fold Like Cross Validation Procedure.}
To evaluate the performance of the hybrid approach in dependence on the size of the training set that is used to train the NBM, we performed several experiments following a k-fold cross-validation like procedure.
However, we slightly adjusted the typical cross-validation procedure to fit our requirements:
From now on, we will refer to the number of training examples in the training set as $n$, where one training example corresponds to one complete observation sequence from the data set.
To evaluate the performance of an approach for a distinct value of $n$, we splitted the complete data set into $k$ partitions, where $k = |\mathcal{D}|/n$ and $|\mathcal{D}|$ is the size of the complete data set (i.e., the number of complete observation sequences).
Then, $k$ models were trained, but in contrast to the typical cross-validation procedure, we always used only \textit{one} of the partitions as the training set and the remaining partitions for validation.
For this procedure, it is important to ensure that the size of the straining set is always equal to $n$.
However, for some values of $n$ and $|\mathcal{D}|$, the data set cannot be splitted into $k$ partitions with equal size.
In such cases, we randomly sampled examples from the other partitions to complete the training set which is constructed from the partition that has a size smaller than $n$.

\textit{Computation of Mean Accuracy.}
All recognition performance results presented in the remainder of this section are reported as the mean accuracy over all recognition problems in the dataset for a relative number of observations $\lambda \in [0, 1]$.
The mean accuracy $Acc$ is calculated as follows:
\begin{equation}
    Acc(\lambda,\mathcal{D}) = \frac{\sum_{R \in \mathcal{D}}{[R(\lfloor T_{R}\lambda\rfloor) = \Tilde{g_R}]}}{|\mathcal{D}|}
\end{equation}
Here, $\mathcal{D}$ is a set of online goal recognition problems $R$, $\Tilde{g_R}$ denotes the correct goal of goal recognition problem $R$, $T_R$ is the maximum value of $t$ for online goal recognition problem $R$ (i.e., length of observation sequence that is associated with $R$), and $[R(t) = \Tilde{g_R}]$ equals 1 if the correct goal is recognized for $R(t)$ and 0 otherwise.
To calculate the reported average accuracy, in contrast to the reported results of \cite{pereira2020landmark}, we only consider the true goal as correctly recognized if the true goal is the \textit{only} goal that is assigned with the maximum heuristic score.

\subsection{Experimental Results and Discussion}
\begin{figure}[t]
    \centering
        \subfloat{
            \begin{tikzpicture}
            \pgfplotsset{every x tick label/.append style={font=\tiny}}
            \pgfplotsset{every y tick label/.append style={font=\tiny}}
                \begin{axis}[
                    width=1\linewidth,
                    height=4.5cm,
                    xlabel={\scriptsize{$\lambda$ (\%)}},
                    ylabel={\scriptsize{$Acc(\lambda, CMU)$}},
                    xlabel near ticks,
                    ylabel near ticks,
                    xmin=0, xmax=24,
                    ymin=0, ymax=1.1,
                    xtick={0,1,2,3,4,5,6,7,8,9,10,11,12,13,14,15,16,17,18,19,20,21,22,23},
                    xticklabels={0,1,2,3,4,5,10,15,20,25,30,35,40,45,50,55,60,65,70,75,80,85,90,95},
                    ytick={0.2,0.4,0.6,0.8,1},
                    legend pos=north west,
                    ymajorgrids=false,
                    xmajorgrids=false,
                    major grid style={line width=.1pt,draw=gray!50},
                    x axis line style={draw=black!60},
                    tick style={draw=black!60},
                    legend columns=4,
                    legend style={draw=none},
                    legend entries={\footnotesize{$PLR_c$},\footnotesize{$PLR_c$},\footnotesize{$PLR_u$},\footnotesize{$PLR_u$}},
                    legend to name={plotLabel3}
                ]
                \addplot[
					color=orange,
					mark=none,
					densely dotted,
					thick,
					]
					coordinates { (1.0,0.324)(2.0,0.324)(3.0,0.372)(4.0,0.473)(5.0,0.5)(6.0,0.507)(7.0,0.547)(8.0,0.676)(9.0,0.851)(10.0,0.932)(11.0,0.966)(12.0,0.966)(13.0,0.993)(14.0,0.993)(15.0,0.993)(16.0,1.0)(17.0,1.0)(18.0,1.0)(19.0,1.0)(20.0,1.0)(21.0,1.0)(22.0,1.0)(23.0,1.0)(24.0,1.0)(25.0,1.0)};
				\addplot[
					color=black,
					mark=none,
					dashed,
					thick,
					]
					coordinates { (1.0,0.311)(2.0,0.311)(3.0,0.297)(4.0,0.291)(5.0,0.297)(6.0,0.297)(7.0,0.297)(8.0,0.324)(9.0,0.372)(10.0,0.459)(11.0,0.507)(12.0,0.534)(13.0,0.581)(14.0,0.595)(15.0,0.635)(16.0,0.703)(17.0,0.804)(18.0,0.838)(19.0,0.872)(20.0,0.878)(21.0,0.899)(22.0,0.932)(23.0,0.959)(24.0,0.98)(25.0,0.98)};
				\addplot[
					color=green,
					mark=none,
					densely dotted,
					thick,
					]
					coordinates { (1.0,0.142)(2.0,0.149)(3.0,0.182)(4.0,0.209)(5.0,0.27)(6.0,0.331)(7.0,0.405)(8.0,0.547)(9.0,0.655)(10.0,0.764)(11.0,0.838)(12.0,0.892)(13.0,0.932)(14.0,0.98)(15.0,0.986)(16.0,0.986)(17.0,1.0)(18.0,1.0)(19.0,1.0)(20.0,1.0)(21.0,1.0)(22.0,1.0)(23.0,1.0)(24.0,1.0)(25.0,1.0)};
				\addplot[
					color=black!50,
					mark=none,
					dashed,
					thick,
					]
					coordinates { (1.0,0.27)(2.0,0.257)(3.0,0.243)(4.0,0.25)(5.0,0.243)(6.0,0.23)(7.0,0.196)(8.0,0.243)(9.0,0.412)(10.0,0.493)(11.0,0.622)(12.0,0.709)(13.0,0.784)(14.0,0.845)(15.0,0.838)(16.0,0.851)(17.0,0.872)(18.0,0.858)(19.0,0.872)(20.0,0.919)(21.0,0.912)(22.0,0.926)(23.0,0.926)(24.0,0.973)(25.0,1.0)};

                \end{axis}
            \end{tikzpicture}}
            \newline
            \vspace{0.2cm}
    \ref{plotLabel3}\vspace{-0.3cm}
    \caption{Comparison of the average recognition accuracy on the CMU Kitchen Dataset for the PLR approach when two different heuristics (i.e., goal completion ($PLR_c$) and uniqueness ($PLR_u$)) are used for goal recognition and initial state landmarks are ignored (dotted lines). Additionally, the performance of the goal completion heuristic is shown for the case when initial state landmarks \textit{are} used (dashed line).}
    \label{fig:initLandmarkComparison}
\end{figure}
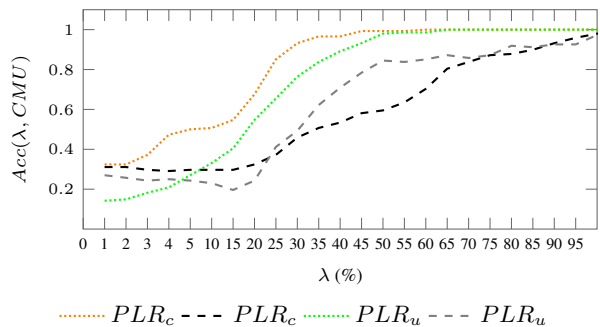
\textit{Investigating the Effect of Ignoring Initial State Landmarks.}
Figure \ref{fig:initLandmarkComparison} shows the average recognition accuracy of the PLR approach on the CMU Kitchen Dataset for both heuristics when no initial state landmarks are used to compute the achieved landmarks (dotted lines).
In addition, the recognition performance when initial state landmarks are used for this computation is depicted (dashed lines).
The results show that the goal completion heuristic clearly outperforms the uniqueness heuristic on the CMU Kitchen dataset.
The major reason for this is most probably that all goals share many landmarks especially early in the observation sequences.
As these goals have not many landmarks with high uniqueness scores, they might be always undervalued compared to other goals that have several landmarks with higher uniqueness scores.
This leads to an overestimation of goals for which the uniqueness scores are very evenly distributed over all landmarks.
As a consequence, when the uniqueness heuristic is used on the CMU Kitchen Dataset, the PLR approach requires a larger fraction of observations to estimate the correct goal compared to when the goal completion heuristic is used.
In addition, the results show that when no initial state landmarks are used, the performance is significantly better than when initial state landmarks are used for both heuristics.
The main reason for this, as already discussed in Section \ref{subsec:computingAchievedLandmarks}, is that initial state landmarks provide no information regarding the most probable goal of an agent that can be derived from observed agent behaviour.

As these results show, that the goal completion heuristic performs much better on the CMU Kitchen Dataset, all experimental results that are presented hereafter do only report the recognition performance for the goal completion heuristic without using initial state landmarks.

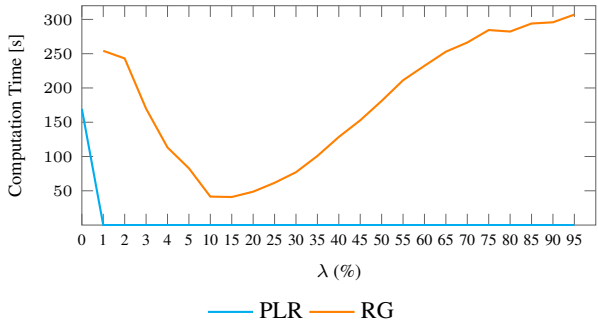
\begin{figure}[h!]
    \centering
        \subfloat{
            \begin{tikzpicture}
            \pgfplotsset{every x tick label/.append style={font=\tiny}}
            \pgfplotsset{every y tick label/.append style={font=\tiny}}
                \begin{axis}[
                    width=1\linewidth,
                    height=4.5cm,
                    xlabel={\scriptsize{$\lambda$ (\%)}},
                    ylabel={\scriptsize{Computation Time [s]}},
                    xlabel near ticks,
                    ylabel near ticks,
                    xmin=0, xmax=24,
                    ymin=0, ymax=320,
                    xtick={0,1,2,3,4,5,6,7,8,9,10,11,12,13,14,15,16,17,18,19,20,21, 22, 23},
                    xticklabels={0,1,2,3,4,5,10,15,20,25,30,35,40,45,50,55,60,65,70,75,80,85,90,95},
                    ytick={50, 100, 150, 200, 250, 300},
                    legend pos=north west,
                    ymajorgrids=false,
                    xmajorgrids=false,
                    major grid style={line width=.1pt,draw=gray!50},
                    x axis line style={draw=black!60},
                    tick style={draw=black!60},
                    legend columns=2,
                    legend style={draw=none},
                    legend entries={\footnotesize{PLR}, \footnotesize{RG}},
                    legend to name={plotLabel2}
                ]

                \addplot[
                    color=cyan,
	                mark=none,
	                thick
	               ]
	                coordinates { (0.0, 169.5)(1.0,0.008)(2.0,0.008)(3.0,0.008)(4.0,0.008)(5.0,0.008)(6.0,0.008)(7.0,0.008)(8.0,0.008)(9.0,0.009)(10.0,0.009)(11.0,0.009)(12.0,0.009)(13.0,0.009)(14.0,0.009)(15.0,0.009)(16.0,0.009)(17.0,0.009)(18.0,0.009)(19.0,0.01)(20.0,0.01)(21.0,0.01)(22.0,0.01)(23.0,0.01)};
	            \addplot[
	                color=orange,
	                mark=none,
	                thick
	                ]
	                coordinates { (1.0,254.2)(2.0,243.1)(3.0,169.8)(4.0,113.2)(5.0,82.8)(6.0,41.5)(7.0,40.9)(8.0,48.7)(9.0,61.6)(10.0,77.1)(11.0,100.8)(12.0,128.5)(13.0,152.8)(14.0,181.1)(15.0,211.2)(16.0,232.3)(17.0,252.9)(18.0,266.3)(19.0,284.5)(20.0,282.3)(21.0,294)(22.0,295.8)(23.0,307)};

                \end{axis}
            \end{tikzpicture}}
            \newline
            \vspace{-0.2cm}
    \ref{plotLabel2}\vspace{0.2cm}
    \caption{Comparison of average required computation time in seconds between the planning landmark based approach (PLR) and the so far used planning based approach (RG) for online goal recognition per goal $g \in G$ over different relative numbers of observation steps.}
    \label{fig:compTimeEvaluation}
\end{figure}
\textit{Comparison of Required Computation Time.}
Figure \ref{fig:compTimeEvaluation} shows the average computation time in seconds that is required per goal $g \in G$ when solving the online goal recognition problems in the CMU Kitchen Dataset.
The computation time is displayed for different relative sizes of the observation sequence.
This means that when an entry in the figure is located at point (1\%, 254.2), the average required computation time to solve the corresponding goal recognition problem that uses an observation sequence that contains the first 1\% of observations of the complete observation sequence was 254.2 seconds.
The results show that the PLR approach requires significantly less computation time than the RG approach, especially once the landmarks were extracted which happens only for $t=0$ in online recognition and takes on average 169.5 seconds.
Once the landmarks were extracted, the PLR approach only requires around 10ms to compute the heuristic score for each goal for all $t \geq 1$.
In contrast, the RG approach requires high amounts of computation time for all values of $t$.
The reason for this is that it does not reuse already computed information when solving an online goal recognition problem but solves $2|G|$ planning problems for each value of $t$.
Interesting to note is also that the RG approach requires rather high amounts of computation time for small values of $t$ and large values of $t$.
The main reason for this is the transformation of the planning domain that the RG approach uses to ensure that the resulting plans fulfill or respectively not fulfill the observation sequence $\pmb{o}$.

\input{figures/figures_tex/evaluation_Landmarks_Hybrid_extended}
In summary, the results show that using the PLR approach instead of the RG approach dramatically reduces the amount of computation time that is required to solve an online goal recognition problem.
As a consequence, the PLR approach enables the hybrid goal recognition method to produce goal probability estimates in an online recognition scenario in near real-time, once the landmarks are extracted.
It is also important to note again that the displayed times are \textit{per potential goal} of the observed agent.
Hence, the PLR approach also scales much better in terms of required computation time when larger sets of potential goals are used.

\textit{Evaluating Hybrid Goal Recognition Performance.}
Figure \ref{fig:landmarkHybridExtended} shows the average accuracy for different sizes of the training data set that is used to train the NBM (i.e., $n$) on the CMU Kitchen Dataset.
The results are displayed for the PLR, RG, and NBM approaches as well as for the hybrid method (HPLR) that uses the PLR method instead of the RG approach.
The results show that the PLR approach consistently outperforms the RG approach when more than 25\% of the observations were seen.
The main reason for the decrease in performance of the RG approach for a relative number of observations larger than 25\% is that the used planner timed out for a large fraction of the involved planning problems.
This shows, once again, that the computation time required by the RG approach is a significant issue when applied to increasingly complex recognition scenarios.
Further, the results show that the hybrid method constantly outperforms or performs similarly well as the two single methods (i.e., PLR and NBM).
Especially when between 5\% and 35\% of the observations are used for the prediction, the hybrid method significantly outperforms both single approaches.
Also interesting to note is that already for rather small training set sizes $n \geq 3$, the NBM always outperforms the PLR for low values of $t$ and the PLR always outperforms the NBM for high values of $t$.
Hence, as the hybrid method always performs at least equally well as the best of the two single approaches, the hybrid method clearly outperforms the PLR approach for low values of $t$ because it can leverage on the strength which the NBM achieves in this area even for small amounts of training data.
Similarly, it clearly outperforms the NBM approach for large values of $t$ because it can leverage on the strength of the PLR approach which performs much better than the NBM approach in this area.
Thus, the hybrid method is able to leverage on the strengths of both single methods and achieves a performance that is superior compared to the performance of both single methods.

\section{Related Work}
\label{sec:relatedWork}
Existing approaches to goal- and plan recognition can be divided into model-based and model-free approaches.
Model-based approaches typically reason over handcrafted symbolic domain models to solve the recognition task.
In contrast, model-free approaches consider the recognition problem as a classification problem and learn to predict the current user goal from data and, thus, are data-driven.

Early model-based approaches to plan recognition relied on complete plan libraries that encode possible user behavior to recognize the current plan from observed user actions \cite{kautz1986generalized,charniak1993bayesian}.
However, these approaches require a large manual modeling effort, which is infeasible in large domains.
To overcome this issue, a new class of approaches to plan recognition that no longer require complete plan libraries, but only a domain model that defines possible states and actions, was proposed.
The symbolic approaches considered in this work (i.e., RG and PLR) \cite{ramirez2010probabilistic},\cite{pereira2020landmark} belong to this class.
Another example approach, that relies on the use of classical planning systems, is the approach by \cite{sohrabi2016revisited}.
They propose to use a top-k planner to generate the top-k plans for all possible goals in order to obtain which goal an observed agent currently intents to achieve.
Nevertheless, most of these approaches have, so far, only been evaluated on relatively small, artificial domains, and hence, it is not clear whether they are also applicable to real-world scenarios.
Moreover, it was recently shown that these approaches have some problems in capturing relations between observations and user goals that cannot be properly modeled manually \cite{wilken2021hybrid}.

In contrast, model-free approaches do not need a domain model but learn from data how to recognize the most probable user goal directly from an observation sequence.
Hence, they have the potential to learn the relations between actions and user goals that are not properly captured by model-based approaches.
In \cite{jameson_towards_1997}, the authors propose to use a BN model to predict the current quest of an observed player of a computer game.
Recently, also some approaches that applied deep learning methods to goal recognition problems appeared \cite{min2016player}, \cite{amado2018lstm}.
For example, \cite{min2016player} applied a LSTM for player goal recognition in digital games.
However, model-free approaches usually require large amounts of training data, which are usually not easily available for real-world scenarios, to produce reasonable results.
Regarding this aspect, model-based approaches have a clear advantage because they can rely on handcrafted domain knowledge.

\section{Conclusion}
\label{sec:conclusion}
In this work we showed that the PLR approach does not only require dramatically less computation time to solve an online goal recognition problem than the originally used RG approach but also significantly improves the goal recognition performance when applied to a real-world goal recognition scenario.
In fact, the PLR approach, once the planning landmarks were initially computed, is able to perform near real-time goal recognition.
Consequently, as the computation time requirements of the RG approach was the major limitation in this regard of the state-of-the-art hybrid method, a PLR based hybrid goal recognition method is also able to perform near real-time goal recognition.
This dramatically improves the applicability of a hybrid goal recognition method in goal recognition scenarios in which reasoning quickly about possible goals is important.
Moreover, the results showed that, similarly to the RG method, the PLR approach has more difficulties to recognize the correct goal early in an observation sequence than a data-driven NBM.
Hence, a hybrid method, which combines the PLR approach and a data-driven NBM, is able to recognize the correct goal more reliably based on a fewer number of observations than the two single approaches.
Nevertheless, we still see some potential to improve the proposed landmark based hybrid recognition approach in future work.
As already mentioned, one limitation of the adjusted landmark extraction algorithm is that ordering information between landmarks is lost.
Investigating this issue is an important path for future work.

\section*{Acknowledgments}
The data used in this paper was obtained from kitchen.cs.cmu.edu and the data collection was funded in part by the National Science Foundation under Grant No. EEEC-0540865.

\bibliographystyle{aaai}
\bibliography{SPARK2022}

\end{document}